\documentclass{article}
\usepackage{spconf,amsmath,graphicx}
\usepackage{amsmath,amssymb}
\usepackage{algorithm}
\usepackage{algpseudocode}
\usepackage{mathrsfs}
\usepackage{amsthm}

\title{Analysis Dictionary Learning: An Efficient and Discriminative Solution}
\name{Wen Tang,\quad Ashkan Panahi, \quad Hamid Krim, \quad Liyi Dai$^\dagger$ 
\sthanks{This research work was generously supported in part by the U.S. Army Research Office under agreement W911NF-16-2-0005.}
}
\address{Department of Electrical and Computer Engineering, North Carolina State University, Raleigh, NC, USA\\
	$^\dagger$Raytheon Integrated Defense Systems, Tewksbury, MA, USA\\
	{\tt\small \{wtang6, apanahi, ahk\}@ncsu.edu, liyi.dai@raytheon.com}}
\begin{document}
%
\maketitle
	\begin{abstract}		
	Discriminative Dictionary Learning (DL) methods have been widely advocated for image classification problems. To further sharpen their discriminative capabilities, most state-of-the-art DL methods have additional constraints included in the learning stages. These various constraints, however, lead to additional computational complexity. We hence propose an efficient Discriminative Convolutional Analysis Dictionary Learning (DCADL) method, as a lower cost Discriminative DL framework, to both characterize the image structures and refine the interclass structure representations. The proposed DCADL jointly learns a convolutional analysis dictionary and a universal classifier, while greatly reducing the time complexity in both training and testing phases, and achieving a competitive accuracy, thus demonstrating great performance in many experiments with standard databases.
\end{abstract}

\section{Introduction}
In the past decade, sparse representation has been widely invoked in many contexts and has been successfully applied to address a variety of image processing and computer vision problems \cite{ksvd,src}. It generally aims to represent data by a linear combination of a few atoms chosen from a data-driven dictionary. 
To pursue such a sparse representation for a particular signal dataset, one well known approach is the Synthesis Dictionary Learning (SDL) \cite{mairal2009non,mairal2009online}, which recovers the signal by learning a dictionary with corresponding coefficients.
SDL can capture complex local structures of images and yield state-of-the-art performance in many image processing problems. Moreover, to overcome the shortcomings of classical patch-based sparse representation and better translation invariance, convolutional filters were also introduced in SDL for signal and image processing applications \cite{szlam2010convolutional,papyan2017convolutional,garcia2018convolutional}. Due to this success in image processing, SDL has also been explored in image inference problems, such as image classification\cite{mairal2009supervised,src,lcksvd,FDDL}, by augmenting with some supervised learning constraints, thereby enhancing the discriminative ability of the resulting dictionaries or sparse representations. In \cite{lcksvd}, Jiang et al. introduced a consistent label constraint together with a universal linear classifier to enforce similarity among representations within the same class. Yang et al. \cite{FDDL} used Fisher Information criterion in their class-specific reconstruction errors to compose their approach.

Besides SDL, Analysis Dictionary Learning (ADL) \cite{analysisksvd,xiao16} has recently been of interest on account of its fast encoding and stability attributes. ADL provides a linear transformation of a signal to a nearly sparse representation. 
Inspired by the SDL methodology in image classification, ADL has also been adapted to the supervised learning problems by promoting discriminative sparse representations \cite{dadl,sadltw}. In \cite{dadl} , Guo et al. incorporated both a topological structure and a representation similarity constraint to encourage a suitable class-selective representation for a $1$-Nearest Neighbor classifier. Tang et al. \cite{tang2018analysis} transformed the original sparse representations with refined and discriminative properties achieved by a jointly learned linear classifier to yield a Structured ADL (SADL) scheme.

In all above methods, both the structure of images and the structure between different classes play important roles in the classification task. Such structures increase the accuracy, but they also require a substantial amount of computation and time for training and testing. It is hence desirable to forego this potentially costly structure-promoting regularization and to instead embed the discriminating characteristics of ADL methods in the dictionary formulation itself.  To this end, we introduce a convolutional mapping within the ADL framework, and embed its resulting feature resolution using its translation invariant structure. We thus propose the Discriminative Convolutional ADL (DCADL) method, which amounts to jointly learning a convolutional ADL and a linear classifier to ensure the capability of characterizing structures among individual images and across classes, while taking advantage of fast ADL encoding. To reduce the excessive training time, we propose a novel algorithmic technique which transforms convolution to a low-cost matrix multiplication. This turns DCADL into an efficiently solvable conventional discriminative ADL framework. 


In Section \ref{sec:ddl}, we describe the generic Discriminative DL framework, followed by the DCADL framework and efficient solution detailed in Section \ref{sec:dcadl}. 
In Section \ref{sec:experiments}, we 
validate our algorithm with standard databases, prior to the conclusions remarks in Section \ref{sec:conclusion}.


\section{Discriminative Dictionary Learning} \label{sec:ddl}
\subsection{Notation}
In this paper, uppercase and lowercase letters respectively denote matrix and vectors throughout the paper. The transpose and inverse of a matrix are respectively denoted by the superscripts $T$ and $-1$. The identity matrix is denoted by $I$.

\subsection{Discriminative Dictionary Learning}
Let $X=[x_1,\dots,x_n] $ denote a training data matrix of $C$ classes and $\Omega $ be an associated dictionary. The conventional Discriminative Dictionary Learning (DL) generally aims to learn efficient and distinct sparse representations $U$ by using feedback from label information. The state-of-the-art Discriminative DL methods \cite{lcksvd,FDDL,dadl,tang2018analysis} generally belong to the following optimization framework,

\begin{equation}\label{equ:ddl}
\resizebox{0.9\columnwidth}{!}{$
\begin{split}
&\arg \min_{\Omega,U}  ~f(\Omega,U,X)+\lambda\|U\|_p + \Phi_S(\Omega,U,Y) +\Phi_G(\Omega,U,Y,W),\\
\end{split}
$}
\end{equation}
with $f(\Omega,U,X)=\frac{1}{2}\|\Omega X-U\|_F^2  ~ \text{ or } ~f(\Omega,U,X)=\frac{1}{2}\|X-\Omega U\|_F^2$, respectively corresponding to ADL and SDL. Furthermore, $\lambda>0$ is a hyper-parameter, $\|\cdot\|_p$ is either $l_0$ or $l_1$ norm to ensure the sparsity of $U$, and $W$ is a classifier. Finally, $Y \in \mathbb{R}^{C \times n}$ represents the labels of the training data, where $Y_{ij}=1$ if and only if image $j$ belongs to class $i$. The dictionary $\Omega$ and the resulting sparse data representation $U$ are jointly learned and adapted for a higher discriminative power by some structure-promoting constraint function $\Phi_S(\Omega,U,Y)$ with a general classification objective functional $\Phi_G(\Omega,U,Y,W)$. 

In this paper, we simplify the Discriminative DL framework in Eq. (\ref{equ:ddl}) to the following by replacing $\Phi_S(\Omega,U,Y)$ by matrix reshaping operators:
\begin{equation} \label{equ:fast-ddl}
\begin{split}
\arg \min_{\Omega,U}  &~f(\Omega,U,X) +\lambda\|\hat{U}\|_p+\Phi_G(\Omega,\tilde{U},Y)\\
s.t. &~ \hat{U}=RS_1(U); ~\tilde{U}=RS_2(U),
\end{split}
\end{equation}
where $RS_1,RS_2$ are some matrix reshaping operators. In order to avoid the direct convolutional computation and expensive costs of updating $\Phi_S(\Omega,U,Y)$ in each iteration, Eq. (\ref{equ:fast-ddl}) significantly improves the DCADL efficiency in both training and testing phases, while maintaining a high classification accuracy, as later substantiated in Section \ref{sec:experiments}.

\section{Discriminative Convolutional Analysis Dictionary Learning}
\label{sec:dcadl}
For clarity, we first formulate an intuitive DCADL framework, and later in Section \ref{sec:efficientDCADL}, and rewrite it to match the structure of that in Eq. (\ref{equ:fast-ddl}). This intuitive DCADL framework is defined as follows,
\begin{equation}\label{equ:dcadl}
\resizebox{0.9\columnwidth}{!}{
$\begin{split}
&\arg \min_{\omega_i,u^i_j,W} 	 ~\sum_{j=1}^n \sum_{i=1}^{m} \left(\frac{1}{2}\|\omega_i \ast x_j-u^i_j\|^2_2 + \lambda_1\|u^i_j\|_1\right)\\
&+\frac{\lambda_2}{2} \|Y-W\tilde{U}\|_F^2,\\
&s.t. ~\|\omega_i\|^2_2 \leq 1; ~\forall i=1,\dots,m, ~\tilde{U}=
\begin{bmatrix}
u^1_1& \cdots & u^1_n \\
\vdots& \ddots & \vdots\\
u^m_1 &\cdots & u^m_n \\
\end{bmatrix},
\end{split}
$}
\end{equation}
where $*$ is convolutional operator, $\omega_i^T \in \mathbb{R}^{s^2}$ is the $i^{\text{th}}$ atom (row) of size $s \times s$ in the analysis dictionary $\Omega$, $x_j \in \mathbb{R}^r$ is the $j^{\text{th}}$ image, and $u^i_j \in \mathbb{R}^p$ is the $i^{\text{th}}$ response map of the $j^{\text{th}}$ image corresponding to the convolution of the $i^{\text{th}}$ atom. Similarly to Eq. (\ref{equ:fast-ddl}), $Y \in \mathbb{R}^{C \times n}$ is the label matrix of training images, and $W \in \mathbb{R}^{C \times mp}$ is the associated linear classifier.

To elaborate on the underlying principle in Eq. (\ref{equ:dcadl}), note that this optimization leads to a set of 2D linear shift-invariant filters, represented by the vectors $\omega_i$ , producing response maps $\omega_i*x_j$ from the images $x_j$. The response maps are nearly sparse in the sense that they possess a suitable sparse approximation given by the vectors $u^i_j$. Furthermore, the response maps are in turn fed to a linear classifier to generate correct labels in $Y$. Imposing sparsity on the response maps provides a better preservation of distinct and valuable information for class-discrimination. Also, note that each image point in the image space is expanded into a high-dimensional vector in the response-map space. In such a response-map space, a one-against-all classifier is also jointly learned to explore the label information and guide the interclass structure of representations. We observe that DCADL and a one-layer Convolutioanl Neural Networks (CNN) exploit similar principles for extracting relevant class-specific information. However, one-layer CNN alternatingly minimizes the first convolutional term and the second classification term, while our algorithm jointly learns these two terms. We omit a more careful discussion in favor of space.

\subsection{Discriminative Convolutional Analysis Dictionary Learning}\label{sec:efficientDCADL}
Noting that conventional ADL formulations rely on matrix multiplication (such as Eq. (\ref{equ:fast-ddl})) for efficient solution, we reformulate our convolutional ADL problem in Eq. (\ref{equ:dcadl}) to be solved in a similar way by assuming that images have no zero-padding. In this case, we segment an image $x_i$ into $p$ patches $[x_{i_1},\dots,x_{i_p}]$ with $s\times s$ pixels, being of the same size as the atom, and let 
$\bar{X}=[x_{1_1},\dots,x_{1_p},\dots,x_{n_1},\dots,x_{n_p}] \in \mathbb{R}^{s^2 \times np}$ and \[\bar{U}=\begin{bmatrix}
u_{1_1}^1&\cdots&u_{1_p}^1&\cdots&u_{n_1}^1&\cdots&u_{n_p}^1\\
\vdots & \ddots &\vdots & \ddots&\vdots &\ddots &\vdots\\
u_{1_1}^m&\cdots&u_{1_p}^m&\cdots&u_{n_1}^m&\cdots&u_{n_p}^m\\
\end{bmatrix} \in \mathbb{R}^{m \times np}.\]
The problem in Eq. (\ref{equ:dcadl}) can then be rewritten in the same form as in Eq. (\ref{equ:fast-ddl}):
\begin{equation}\label{equ:dcadlp}
\resizebox{0.9\columnwidth}{!}{
$\begin{split}
&\arg \min_{\substack{\Omega, \bar{U}, W \\ \hat{U}, \tilde{U}}} 	 ~\frac{1}{2}\|\Omega \bar{X}-\bar{U}\|^2_F + \lambda_1\|\hat{U}\|_1+\frac{\lambda_2}{2}  \|Y-W\tilde{U}\|_F^2,\\
&s.t.  ~\|\omega_i\|^2_2 \leq 1; ~\forall i=1,\dots,m,\\
&~\hat{U}=
\begin{bmatrix}
u^1_{1_1}& u^2_{1_1}&\cdots & u^m_{n_1} \\
\vdots& \vdots& \ddots & \vdots\\
u^1_{1_p} &u^2_{1_p} &\cdots & u^m_{n_p} \\
\end{bmatrix}, ~\tilde{U}=
\begin{bmatrix}
u^1_{1_1}& \cdots & u^1_{n_1} \\
\vdots& \ddots & \vdots\\
u^m_{1_p} &\cdots & u^m_{n_p} \\
\end{bmatrix}.
\end{split}
$
}
\end{equation}
It is noteworthy that $\bar{U} \in \mathbb{R}^{m \times np}, \hat{U} \in \mathbb{R}^{p \times mn}$ and $\tilde{U} \in \mathbb{R}^{mp \times n}$ are merely different reshapings of $U=[u^i_{j_k}]\in \mathbb{R}^{p \times n \times m}, \forall i=1,\dots,m,~\forall j=1,\dots,n,~\forall k=1,\dots,p$, where $u^i_{j_k}$ is the vectorized response map of the $i^{\text{th}}$ atom and the $p^{\text{th}}$ patch of the $j^{\text{th}}$ image.

\subsection{Algorithmic Solution}\label{sec:algorithm}
Although our optimization problem in Eq. (\ref{equ:dcadlp}) is non-covex, it is still a multi-convex problem. Therefore, we may reliably update the variables by the block-coordinate descent method. 
We follow the updating steps in each iteration of our algorithm, which are summarized in Algorithm 1.
\renewcommand{\algorithmicrequire}{\textbf{Input:}}
\renewcommand{\algorithmicensure}{\textbf{Output:}}
\begin{algorithm}[htb] \label{alg:learning}  
	\caption{DCADL}  
	\begin{algorithmic}[1] 
		\Require 
		Training data $\bar{X}=[x_{1_1},\dots,x_{n_p}]$, classes labels $Y$, parameter $\lambda_1$, $\lambda_2$, $\lambda_3$, $\lambda_4$, and maximum iteration $T$;
		\Ensure $\Omega$,  ${U}$, and $W$;
		\State Initialize $\Omega$, $U$, and $W$;
		\While {not converged \textbf{and} $t < T$}
		\State t=t+1; {\scriptsize \% $\rho$ is learning rate.}
		\State Update $\bar{U}_{t+1}$ 
		by $\bar{U}_{t+1}=\bar{U}_{t}-\rho (\bar{U}_t-\Omega \bar{X});$
		\State $\tilde{U}_t=R_1(\bar{U}_{t+1});$ {\scriptsize\%$R_1(\cdot)$ is reshaping operator.}
		\State Update $\tilde{U}_{t+1}$ 
		by $\tilde{U}_{t+1}=\tilde{U}_{t}-\rho (\lambda_2 W^T(L-W\tilde{U}_t));$
		\State $\hat{U}_t=R_2(\tilde{U}_{t+1});$ {\scriptsize\%$R_2(\cdot)$ is reshaping operator.}
		\State Update $\hat{U}_{t+1} $
		by $\hat{U}_{t+1}=\tau_{\rho \lambda_1}(\hat{U}_t);$
		\State $\tilde{U}_{t+1}=R_2^{-1}(\hat{U}_{t+1})$; 
		\State Update $W_{t}$ by {\footnotesize $W_{t+1}=\lambda_2 Y\tilde{U}_{t+1} (\lambda_2 \tilde{U}_{t+1}\tilde{U}^T_{t+1}+\lambda_3 I)^{-1};$}
		\State $\bar{U}_{t+1}=R_1^{-1}(\tilde{U}_{t+1})$; {\scriptsize \%$R_1^{-1},R_2^{-1}$ is inverse reshaping operator, }
		\State Update $\Omega_{t}$ by $\Omega_{t+1}=\bar{U}_{t+1}\bar{X}^T(\bar{X}\bar{X}^T+\lambda_4 I)^{-1};$
		\State Normalize $\Omega_k$ by $\omega_i^T=\frac{\omega_i^T}{\|\omega_i^T\|_2}, \text{ if } \|\omega_i^T\|^2_2 >1, \forall i$;
		\EndWhile
	\end{algorithmic}  
\end{algorithm}

\section{Experiments and Results}\label{sec:experiments}
Four widely used visual classification datasets, Extended YaleB\cite{yaleB}, AR\cite{AR}, Caltech101\cite{caltech101}, and Scene15\cite{scene15}, have been applied to evaluate our proposed DCADL. 

In our experiments, a comprehensive evaluation with classification accuracy, training time and testing time is provided. 
The testing time is computed by the average processing time to classify a single image.\par

To evaluate our proposed DCADL, we carry out a comparative study with the following methods: The first one is ADL+SVM\cite{shekhar2014analysis}, which serves as a baseline. 
LC-KSVD \cite{lcksvd} is a state-of-the-art SDL. 
Then SADL\cite{tang2018analysis} and DADL\cite{dadl} are up-to-date ADL approaches. The last method, DPL\cite{dpl} is a hybrid technique of SDL and ADL.

The parameters $\lambda_1$, $\lambda_2$, $\lambda_3$, $\lambda_4$ and $T$ are chosen by a 10-fold cross validation on each dataset. The parameters of all competing methods are also optimally tuned to ensure their best performance. 
The different atom numbers employed in each approach will be listed in parentheses in our Tables. Moreover, we show the reported accuracy for the benchmark methods in their original paper in parentheses with the appropriate citation. The difference in the accuracy between our implementation and the originally reported one might be due to different segmentations of the training and testing samples.

\begin{figure}[htb]
	\centering
	\includegraphics[width=0.3\textwidth]{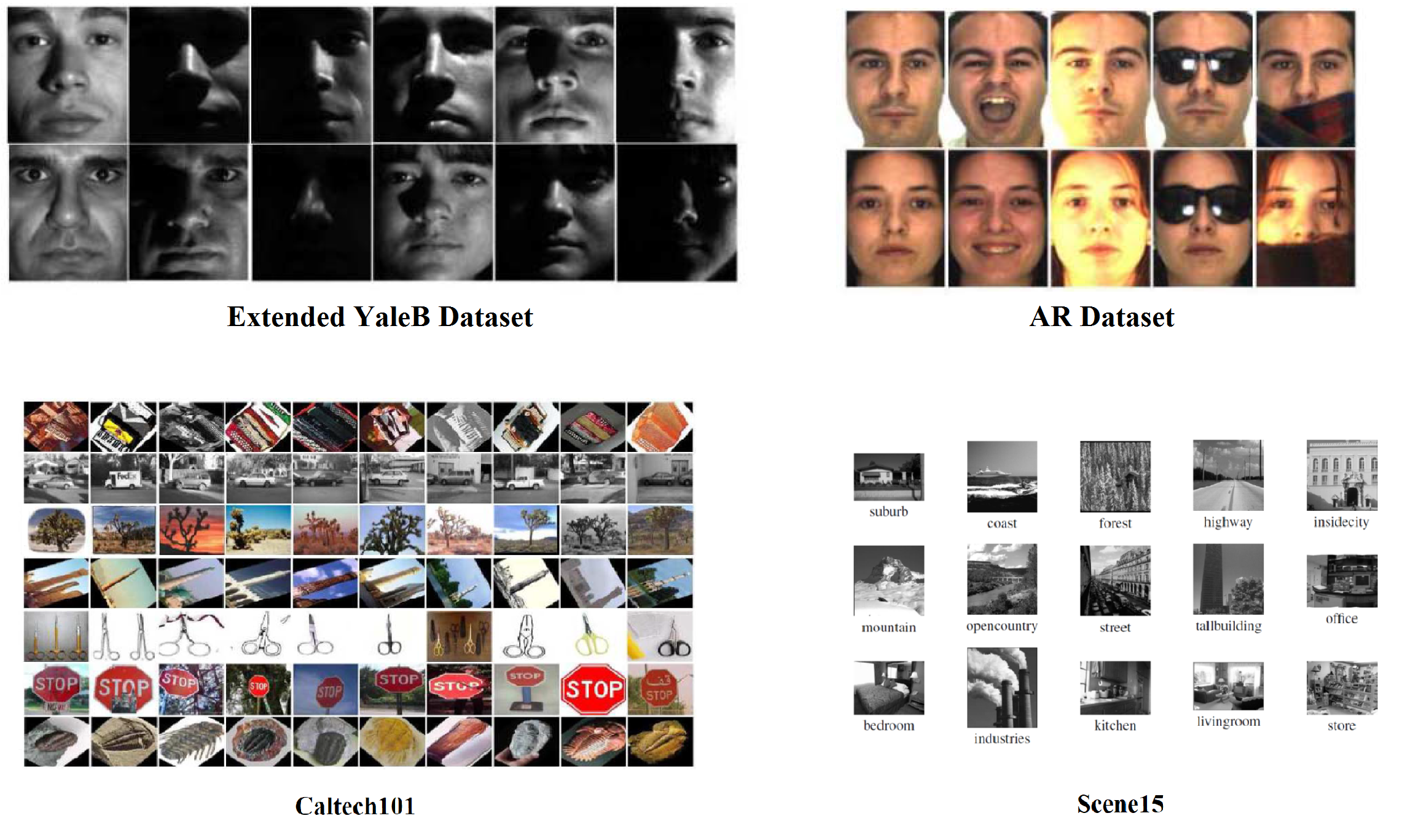}
	\caption{Examples of Four Different Datasets.}
	\label{fig:YaleB}
\end{figure}

\subsection{Extended YaleB}
There are in total 2414 frontal face images of 38 people. 
We cropped each image to $48 \times 42$ pixels as illustrated in the left-up corner of Figure \ref{fig:YaleB}. We randomly choose half of the images for training, and the rest for testing. In our experiment, each analysis atom is $12 \times 12$ pixels and convolves with each image with the stride of 6. The dictionary size of our DCADL is 50 atoms, $\lambda_1=0.001$, $\lambda_2=0.2$, $\lambda_3=0.1$, $\lambda_4=0.1$ and $T=23$.

\begin{table}[htb]
	\centering
	\caption{Classification Results on Extended YaleB Dataset}
	\resizebox{\columnwidth}{!}{%
	
	\begin{tabular}{c c c c}
		\hline
		Methods($\#$atoms) &  Accuracy(\%) &  Training Time(s) &  Testing Time(s)\\
		\hline
		ADL+SVM(1216)\cite{shekhar2014analysis} &  $88.91 \pm 0.73$ &  $274.16$ &  $6.78\times 10^{-4}$\\
		LC-KSVD(570)\cite{lcksvd} &  $94.74\pm 0.47$ &  $183.55$  &  $1.36\times 10^{-3}$\\
		LC-KSVD(1216)\cite{lcksvd}&  $66.05\pm 2.35$ &  $244.77$  &  $1.23\times 10^{-3}$\\
		SADL(1216)\cite{tang2018analysis} &  $97.58\pm 0.39$ &  $257.31$ &  $\bf{1.53\times 10^{-5}}$\\
		DADL(2031)\cite{dadl} &  $98.33\pm0.28$ & $6.40$ &  $2.19 \times 10^{-4}$\\
		DPL(1216)\cite{dpl} &  $98.01\pm 0.45$ &  $20.25$&  $2.09 \times 10^{-4}$\\
		\hline
		HDL-2 (-)\cite{HDL} & $98.50$ &  - &  -\\
		PCANet-1 (-)\cite{PCANET} &  $97.77$ &  -&  -\\
		\hline
		DCADL(50) &  $\bf{99.57 \pm 0.08}$&  $\bf{3.82}$ &  $1.93\times 10^{-5}$ \\
		\hline
	\end{tabular}%
	}
	\label{tab:yaleb}%
\end{table}%

The classification results, training and testing times are summarized in Table \ref{tab:yaleb}. In the second part of Table \ref{tab:yaleb}, a 2-layer hierarchical dictionary learning approach\cite{HDL} and a 1-layer convolutional network\cite{PCANET} are also included for comparison. Both of these two methods also worked on raw pixels of images. Our proposed DCADL method achieves the highest classification accuracy with the shortest training time and an extremely fast testing time, while securing an at least $1\%$ greater accuracy relative to others'.

\subsection{AR Face}

The AR Face dateset consists of 2600 color images of 50 females and 50 males 
. We then cropped each image to $55 \times 40$, which is shown in the right-up corner of Figure \ref{fig:YaleB}. 20 images per class are randomly selected to form a training set, and others are used for testing. Similarly to the settings in Extended YaleB, the convolutional analysis atom size is also $12 \times 12$ pixels with the stride of 6. The dictionary size of our DCADL is 50 atoms, $\lambda_1=0.0001$, $\lambda_2=0.005$, $\lambda_3=0.0001$, $\lambda_4=1.3$ and $T=37$.

\begin{table}[htb]
	\centering
	\caption{Classification Results on AR Dataset}
		\resizebox{\columnwidth}{!}{%
	\begin{tabular}{c c c c}
		\hline
		Methods($\#$atoms) & Accuracy(\%) & Training Time(s) & Testing Time(s)\\
		\hline
		ADL+SVM(2000)\cite{shekhar2014analysis} & $85.35\pm2.34$ & $1301.97$ & $9.05\times 10^{-3}$\\
		LC-KSVD(500)\cite{lcksvd} & $91.97 \pm 1.09 $ & $275.18$  & $3.93\times 10^{-4}$\\
		LC-KSVD(2000)\cite{lcksvd} & $67.70 \pm 5.14 $ & $253.55$  & $2.31\times 10^{-3}$\\
		SADL(2000)\cite{tang2018analysis} & $98.55 \pm 0.33$ & $69.93$ & ${2.88 \times 10^{-5}}$\\
		DADL(2211)\cite{dadl} & $\bf{99.20\pm 0.28}$ & $\textbf{10.42}$ & $4.26\times 10^{-4}$\\
		DPL(2000)\cite{dpl} &$99.03\pm0.32$ &$24.03$ &$8.45\times 10^{-5}$\\
		\hline
		CNN-3 (-)\cite{CNNAR} & $96.50$ & -&-\\
		\hline
		DCADL(50) &$98.93 \pm 0.43$& $14.52$ &$\bf{2.78\times 10^{-5}}$ \\
		\hline
	\end{tabular}%
	}
	\label{tab:ar}%
\end{table}%
The classification results as well as the training and testing times are summarized in Table \ref{tab:ar}. The accuracy of our proposed DCADL is barely lower than DADL and DPL, but it is still much higher than other methods with a very quick training and testing time. It is even better than the performance of a 3-layer Convolutional Network\cite{CNNAR}, which also worked on the raw pixel of the AR dataset. Though DADL has a faster training time than DCADL, its testing time is still 10 times slower than ours, and it needs to calculate a weight matrix in advance to keep its crucial topological structure, which is time consuming.


\subsection{Caltech101}
The Caltech101 dataset includes 101 different object categories and a non-object category, as shown left-down corner of Figure \ref{fig:YaleB}.  
The standard bag-of words+spatial pyramid matching (SPM) framework \cite{scene15} is used to calculate the SPM features. PCA is finally applied to the vectorized SPF to reduce its dimension to $3000$. In our experiment, 30 images per class are randomly chosen as training data, and other images are used as testing data. All above steps and settings follow \cite{lcksvd}. As features are vectors, our DCADL uses 1-dimensional convolution for such features. The convolutional analysis atom size is $1500 \times 1$ with step of $1500$. The dictionary size of our DCADL is 152, $\lambda_1=0.0001$, $\lambda_2=0.01$, $\lambda_3=0.006$, $\lambda_4=0.15$ and $T=48$.

\begin{table}[htb]
	\centering
	\caption{Classification Results on Caltech101 Dataset}
		\resizebox{\columnwidth}{!}{%
	\begin{tabular}{c c c c }
		\hline
		Methods($\#$atoms) & Accuracy(\%) &Training Time(s) & Testing Time(s)\\
		\hline
		ADL+SVM(3060)\cite{shekhar2014analysis} & $66.75\pm 1.08$ & $1943.47$ & $1.33\times 10^{-2}$\\
		LC-KSVD(3060)\cite{lcksvd} & $73.67 \pm 0.93$ (73.6\cite{lcksvd}) & $2144.90$ & $2.49 \times 10^{-3}$\\
		SADL(3060)\cite{tang2018analysis} & $\bf{74.17 \pm 0.49}$ (\cite{tang2018analysis}) & $1406.68$ & $4.76\times 10^{-5}$\\
		DADL(3061)\cite{dadl} &$71.77\pm0.44$ (\bf{74.6}\cite{dadl}) & $26.29$ & $7.90\times 10^{-4}$\\
		DPL(3060)\cite{dpl} &$71.64\pm0.50 $ (73.9\cite{dpl}) &$64.33$ &$3.79\times 10^{-4}$\\
		DCADL(152) &$\bf{74.17 \pm 0.42}$ &$\textbf{17.55}$ &$\bf{2.52\times 10^{-5}}$ \\
		\hline
	\end{tabular}%
	}
	\label{tab:caltech101}%
\end{table}%

The classification results, training and testing times are summarized in Table \ref{tab:caltech101}. DCADL achieves the highest performance again in our experiments, achieving the fastest training and testing time. Though its accuracy is slightly lower than the reported one in DADL\cite{dadl}, DCADL is at least 1.5 times faster than DADL in training and testing time.

\subsection{Scene15}
The Scene15 dataset has 15 different scene categories, which are shown righ-down corner of Figure \ref{fig:YaleB}. 
We extracted the SPM features for Scene 15 dataset by the same procedures as for Caltech 101. 100 images per class are randomly picked as training data, and the rest is used for testing data. The settings and steps also follow \cite{lcksvd}. Similarly to the setting of Caltech101, the convolutional analysis size of DCADL is $1500 \times 1$ with a $1500$ step. The dictionary size is 100, $\lambda_1=0.01$, $\lambda_2=0.5$, $\lambda_3=0.09$, $\lambda_4=0.55$ and $T=15$.

\begin{table}[htb]
	\centering
	\caption{Classification Results on Scene15 Dataset}
		\resizebox{\columnwidth}{!}{%
	\begin{tabular}{c c c c}
		\hline
		Methods($\#$atoms) & Accuracy(\%) & Training Time(s) & Testing Time(s)\\
		\hline
		ADL+SVM(1500)\cite{shekhar2014analysis} & $80.55\pm 3.20$ &$494.41$ & $1.73\times 10^{-4}$\\
		LC-KSVD(1500)\cite{lcksvd} & $\bf{99.21\pm0.18}$ (92.9\cite{lcksvd}) & $390.22$ & $1.81\times 10^{-3}$\\
		SADL(1500) & $98.40\pm0.21$ (-) & $219.80$ &${2.41\times 10^{-5}}$\\
		DADL(3001)\cite{dadl} &$97.81\pm0.27$ (98.3\cite{dadl})&$15.00$ &$4.62\times 10^{-4}$\\
		DPL(1500)\cite{dpl} &$98.35\pm0.17$ (97.7 \cite{dadl}) &$8.83$ &$5.67\times 10^{-5}$\\
		DCADL(50) &$\bf{98.41\pm0.26}$ &$\textbf{2.59}$ & $\bf{1.00\times 10^{-5}}$\\
		\hline
	\end{tabular}%
	}
	\label{tab:scene15}%
\end{table}%

The classification results of each method are summarized in Table \ref{tab:scene15}. Our accuracy is barely lower than LC-KSVD, but is still higher than all other methods and the reported performance in LC-KSVD. In addition, compared with all other methods, DCADL still registers a much greater training and testing time gain.

\section{Conclusions} \label{sec:conclusion}
We proposed an efficient discriminative convolutional ADL method for classification tasks. Our DCADL consists of learning a convolutional ADL together with a universal linear classifier. We further transformed the optimization framework of DCADL to a more efficient discriminative DL framework by eliminating structural constraint costs, while preserving the discriminative power. Our extensive numerical studies show the DCADL exhibits its highly competitive accuracies with significant efficiency.
\nocite{*}

%
%

\small
\bibliographystyle{IEEEbib}
\bibliography{egbib}

\end{document}